\documentclass[runningheads]{llncs}
\usepackage{graphicx}
\usepackage{hyperref}
\begin{document}
%

\title{Kwame for Science: An AI Teaching Assistant Based on Sentence-BERT for Science Education in West Africa}

%
\titlerunning{Kwame for Science}
%
%
\author{George Boateng\inst{1,2} \and
Samuel John\inst{1} \and
Andrew Glago\inst{1}\and
Samuel Boateng \inst{1}\and
Victor Kumbol \inst{1,3}
}

\institute{SuaCode.ai, Inc., U.S. \and ETH Zurich, Switzerland \and Charite Berlin, Germany\\
\email{\{jojo, samuel.john, andrew.glago, samuel.boateng,  victor\}@suacode.ai}
}

\authorrunning{Boateng et al.}
%
%
\maketitle              
\begin{abstract}
Africa has a  high student-to-teacher ratio which limits students’ access to teachers. Consequently, students struggle to get answers to their questions. In this work, we extended Kwame, our previous AI teaching assistant, adapted it for science education, and deployed it as a web app. Kwame for Science answers questions of students based on the Integrated Science subject of the West African Senior Secondary Certificate Examination (WASSCE). Kwame for Science is a Sentence-BERT-based  question-answering web app that displays 3 paragraphs as answers along with a confidence score in response to science questions. Additionally, it displays the top 5 related past exam questions and their answers in addition to the 3 paragraphs. Our preliminary evaluation of the Kwame for Science with a 2.5-week real-world deployment showed a top 3 accuracy of 87.5\% (n=56) with 190 users across 11 countries. Kwame for Science will enable the delivery of scalable, cost-effective, and quality remote education to millions of people across Africa. \footnote{Copyright © 2022 for this paper by its authors. Use permitted under Creative Commons License Attribution 4.0 International (CC BY 4.0).} 

\end{abstract}

\section{Introduction}
The COVID-19 pandemic has exacerbated the already poor educational experiences of millions of students in Africa who were grappling with educational challenges like poor access to computers, the internet, and teachers. In 2018, the average student-teacher ratio in Sub-Saharan Africa was 35:1 which is higher compared to 14:1  in Europe \cite{UNESCO2020}. In this context, students struggle to get answers to their questions. Hence, offering quick and accurate answers, outside of the classroom, could improve their overall learning experience. However, it is difficult to scale this support with human teachers.

In 2020, we developed Kwame \cite{boateng2021b}, a bilingual AI teaching assistant that provides answers to students’ coding questions in English and French for SuaCode, a smartphone-based online coding course \cite{boateng2019,boateng2021}. Kwame is a deep learning-based question answering system that finds the paragraph most semantically similar to the question via cosine similarity with a Sentence-BERT model. We extended Kwame to work for science education and deployed it as a web app. Specifically, Kwame for Science \footnote{\href{http://kwame.ai/}{http://kwame.ai/}} answers questions of students based on the Integrated Science subject of the West African Senior Secondary Certificate Examination (WASSCE). This is a core subject that covers various aspects of science such as biology, chemistry, physics, earth science, and agricultural science. It is mandatory for senior high school students in the West African Education Council (WAEC) member countries (Ghana, Nigeria, Sierra Leone, Liberia, and The Gambia).

There are virtual teaching assistants (TA) such as Jill Watson \cite{goel2016,goel2020}, Rexy \cite{benedetto2019}, and a physics course TA \cite{zylich2020} and Curio SmartChat (for K-12 science) \cite{raamadhurai2019} (see \cite{boateng2020} for a detailed description of related work). These works are focused on answering logistical questions, except Curio SmartChat. In comparison to Curio SmartChat which is the closest work to ours, our work uses a state-of-the-art language model (Sentence-BERT) relative to theirs (Universal Sentence Encoder). Also, our work is the first to be developed and deployed in the context of high school science education in West Africa.

\section{Kwame for Science  System Architecture}
\begin{figure}
\includegraphics[width=\linewidth]{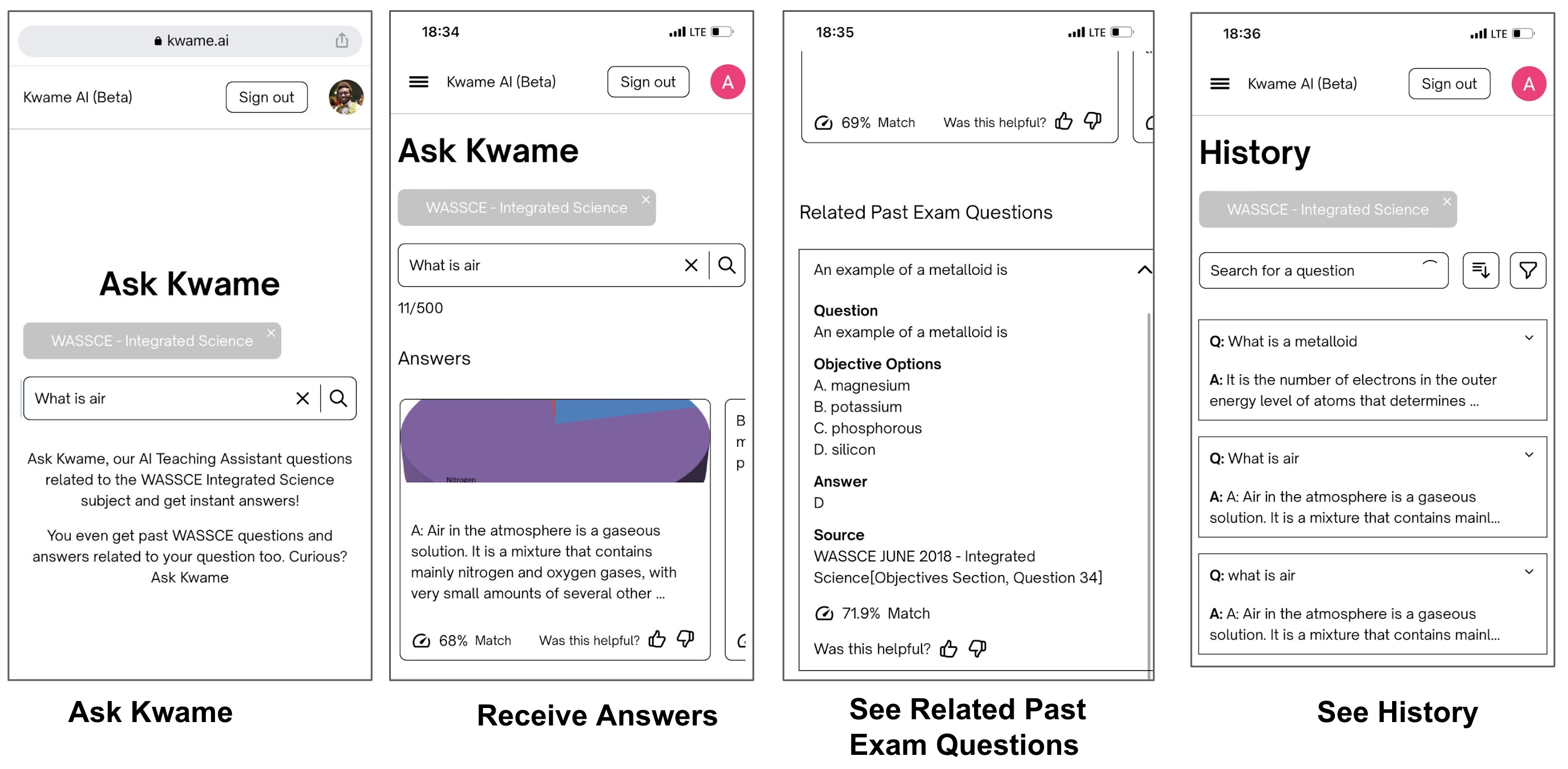}
\caption{Screenshots of Kwame for Science} 
\label{fig:kwame4science}
\end{figure}

Kwame for Science is a Sentence-BERT-based question-answering web app that displays 3 paragraphs as answers along with a confidence score which represents the similarity score in response to science questions (Figure \ref{fig:kwame4science}). Additionally, it displays the top 5 related past exam questions and their answers in addition to the 3 paragraphs. We used a Sentence-BERT (SBERT) model that was pretrained on a large and diverse set of question-answer pairs. We used the SBERT model as it was, with plans for fine-tuning after real-world data collection especially since exploratory evaluation for our science use case showed it had decent performance.

When a user types a question in the web app, our system computes an embedding of the question using the SBERT model. Next, it computes cosine similarity scores with a bank of answers (which are paragraphs from our knowledge source), retrieves, and returns the top 3 answers along with a confidence score and any figures or images referenced in that paragraph to the web app. Additionally, it computes cosine similarity scores with a bank of past exam questions, retrieves, and returns the top 5 related questions and their answers, along with confidence scores. The web app then displays the answers and the related past exam questions that are above a preset similarity score threshold. If no answer is above the threshold, a message is shown saying the question could not be answered using the knowledge source of that subject. We precomputed embeddings for fast real-time retrieval and saved them as indices in ElasticSearch which we hosted on Google Cloud Platform. 

\section{Dataset Curation and Preprocessing}
Given that our goal was for Kwame to provide answers based on the Integrated Science subject of the WASSCE exam, our training data and knowledge source had to cover the topics in the WASSCE Integrated Science curriculum. We sought to use one of the approved textbooks in Ghana. Unfortunately, their copyrights did not permit such use and the publishers were unwilling to partner with us. Consequently, we searched for free and open-source books and datasets that fulfilled our needs. We came across a middle school science dataset — Textbook Questions Answering (TQA) \cite{kembhavi2017} which was curated from the free and open-source textbook, CK-12. Our exploration of the dataset revealed that though it covered several of the WASSCE Integrated Science topics, it lacked others, particularly those related to agricultural science. Consequently, we additionally used a dataset based on Simple Wikipedia to cover those gaps. We used Simple Wikipedia since its explanations were simple and better suited for middle school and high school students compared to regular Wikipedia.

We parsed the JSON files of the dataset into paragraphs. We also extracted figures that were referenced in the paragraphs so they could be returned to students along with the answers. We then split the paragraphs into groups of 3 sentences, computed embeddings, and indexed them using ElasticSearch to enable fast retrieval and run time. These constituted the answers returned for questions. Furthermore, we augmented our question-answering with curriculum-specific content. In particular, we created question-answer pairs using WASSCE questions that cover exams from 2000 to 2020. The exam has three parts, objectives (multiple-choice), theory, and practicals. Similar to the paragraphs, we computed embeddings of the questions and indexed them using ElasticSearch. These constituted the related past questions (with answers) returned when a question is asked.

\section{Preliminary Evaluation and Results}
We launched the web app in beta on 10th June 2022. Users could provide feedback by upvoting or downvoting answers in response to the question “Was this helpful?.” To evaluate Kwame for Science, we used the metrics top 1 and top 3 accuracies. Top 1 accuracy quantifies performance assuming only one answer was returned and voted on. Top 3 accuracy refers to the performance where for each question that received a vote, at least one answer was rated as helpful out of the 3 answers that were returned. The statistics for the deployment between 10th June 2022 and 27th June 2022 (2.5 weeks) are 190 users across 11 countries (6 in Africa), 433 questions with the metrics 71.8\%  top 1 accuracy (n=117 answers), and 87.5\%  top 3 accuracy (n=56 questions). The top 3 accuracy result is good, showing that Kwame for Science has a high chance of giving at least one useful answer among the 3. Some challenging cases occurred when there were typos in the spelling of scientific words and the questions were related to topics outside the scope of the knowledge source. Also, some unhelpful answers were cases where the returned paragraph was incomplete due to issues with the dataset.

\section{Conclusion}
In this work, we developed and evaluated Kwame for Science which provides instant answers to the Science questions of students across West Africa. Our future work will fine-tune the SBERT model using the real-world votes on answers to improve its accuracy. Also, we will make Kwame for Science available in local languages across Africa, and available via offline channels such as SMS, USSD, and toll-free calling. Kwame for Science will enable the delivery of scalable, cost-effective, and quality remote education to millions of people across Africa.

\section{Acknowledgement}
This work was supported with grants from ETH for Development (ETH4D) and the MTEC Foundation, both at ETH Zurich.

%
%
%
\bibliographystyle{splncs04}
\bibliography{refs}
\end{document}